# DEEPFAKE DETECTION VIA FACIAL FEATURE EXTRACTION AND MODELING

Benjamin Carter, Nathan Dilla, Micheal Callahan, Atuhaire Ambala, Grand Canyon University

## ABSTRACT

*The rise of deepfake technology brings forth new questions about the authenticity of various forms of media found online today. Videos and images generated by artificial intelligence (AI) have become increasingly more difficult to differentiate from genuine media, resulting in the need for new models to detect artificially-generated media. While many models have attempted to solve this, most focus on direct image processing, adapting a convolutional neural network (CNN) or a recurrent neural network (RNN) that directly interacts with the video image data. This paper introduces an approach of using solely facial landmarks for deepfake detection. Using a dataset consisting of both deepfake and genuine videos of human faces, this paper describes an approach for extracting facial landmarks for deepfake detection, focusing on identifying subtle inconsistencies in facial movements instead of raw image processing. Experimental results demonstrated that this feature extraction technique is effective in various neural network models, with the same facial landmarks tested on three neural network models, with promising performance metrics indicating its potential for real-world applications. The findings discussed in this paper include RNN and artificial neural network (ANN) models with accuracy between 96% and 93%, respectively, with a CNN model hovering around 78%. This research challenges the assumption that raw image processing is necessary to identify deepfake videos by presenting a facial feature extraction approach compatible with various neural network models while requiring fewer parameters.*

## INTRODUCTION

The impacts of artificial intelligence (AI) in society have had wide-reaching effects, challenging how society interacts with digital media today. One such challenge is the authenticity of images and videos of a given person or event. With the rise of more powerful models in artificial intelligence, these models have become capable of generating realistic video, audio, and images (Ademiluyi, 2023; Karandikar et al., 2020). A *deepfake* has become the term of choice to describe artificially-generated images and video clips of famous people, making it appear that the person did something they never did. Although many deepfakes have been created for entertainment, such as political candidates singing popular songs, deepfakes have also been used to manipulate groups of people to act, think, or believe in a certain way (Diakopoulos & Johnson, 2020). Therefore, finding a method to detect these deepfakes effectively is critical to inform the public on the legitimacy of various images and recordings. Work has been done in building models to detect deepfakes; however, many of these models in use today focus on raw image processing (Güera & Delp, 2018). This paper introduces an alternative feature extraction method that relies on facial landmark data instead of raw image data, aiming to reduce computational demands compared to traditional deepfake detection approaches. This research focuses on analyzing artificial videos of faces. For AI-generated videos featuring humans, analyzing facial movements can provide indicators of manipulation (Li et al., 2024). This suggests that tracking the progression of facial features across frames may offer comparable detection capabilities while lowering the required amount of data for training a model normally found in raw-image processing, therefore optimizing the modeling process.

This facial feature extraction method is then tested across various neural network architectures with successful results, demonstrating that this method offers an alternative to the raw image processing commonly used today. We tested the same extracted dataset in a recurrent neural network (RNN), a convolutional neural network (CNN), and an artificial neural network (ANN). An RNN accuracy resulted in around 96%, a CNN in 78%, and an ANN in 93%. The software for the feature extraction and modeling is available online at our repository (https://zenodo.org/records/15377237).

## RELATED WORK

As deepfake technology rapidly advances, effective detection methods must also improve to keep pace. Current work in the field gravitates toward direct image manipulation. Detection efforts fall into three broad methodological camps: image-based convolutional neural networks (CNNs), hybrid spatiotemporal networks, and landmark-based feature methods. Image-based CNN models operate directly on raw frames to learn pixel-level artifacts (Karandikar et al., 2020). Researchers found that standard CNNs can reliably spot lower-quality deepfakes by picking up on texture and blending inconsistencies. Building on this, Ademiluyi (2023) suggested first applying a face detector to crop facial regions, then passing those regions through a CNN to extract fine-grained feature representations, yielding greater robustness than whole-frame approaches.

Hybrid spatiotemporal networks combine CNNs for spatial cues with recurrent neural networks (RNNs) for temporal consistency. Jamsheed and Janet (2021) used a pre-trained ResNeXt CNN to extract frame-wise facial embeddings, then fed sequences of 10–80 frames into an RNN, reporting strong detection rates. Likewise, Güera and Delp (2018) combined CNN and RNN layers to achieve ~99% training accuracy and ~96% test accuracy. However, given the rapid improvement of generative models since 2018, such architectures may struggle against today's high-fidelity deepfakes and often incur substantial computational costs. Today's generative models involve a series of steps in generating a deepfake image starting frequently with splitting a video into frames (Shah, 2024). Models first extract facial landmarks of a reference image, align the face using the facial landmarks, and then superimpose the edited image on the facial landmarks. Taking this, Li et al. (2024) created a system to obstruct deepfake generation by altering the facial landmarks before the alignment stage in the deepfake pipeline. They observed that altering the facial landmarks obstructs the model from generating quality deepfake outputs. Deepfake generation is therefore, highly reliant on identifying certain facial features for image manipulation.

Our research sought to take this further to find the inconsistencies in the generated image's facial landmarks. Even when a deepfake generation algorithm utilizes a reference image for facial landmarks, the algorithm cannot fully stretch the target image to fit all 60+ landmarks on the face. While modern models mainly rely on raw image processing, we sought to create a facial extraction technique that can capitalize on these subtle inconsistencies to accurately classify artificially generated videos while removing the need for direct pixel analysis.

## PROJECT ARCHITECTURE

The architecture of the project was divided into two components, the feature extraction process and the model testing process. The following sections detail these two components.

## DATA EXTRACTION

This project analyzed a deepfake detection dataset available through Kaggle, which contains videos of individuals in various day-to-day conversations and activities (Tiwarekar, 2024). This dataset featured 364 authentic videos and 3,068 artificially generated videos. For processing, both sets of videos were taken through an algorithm (Figure 1. The first step of the algorithm involved cropping the frame to focus solely on the face (see *a* in Figure 1), similar to Ademiluyi (2023), as this reduces false positives in the next step. The algorithm then analyzed the cropped region to report whether there was a face (see *b* in Figure 1). It used an OpenCV Haar Cascade classifier to identify the existence of a face, as this effectively finds objects in an image (Singh et al., 2013). If a face was found, it ran another Haar Cascade classifier to find an eye in the face. This second pass in finding an eye protects against further false positives.

*Figure 1.*
*Feature Extraction Architecture*

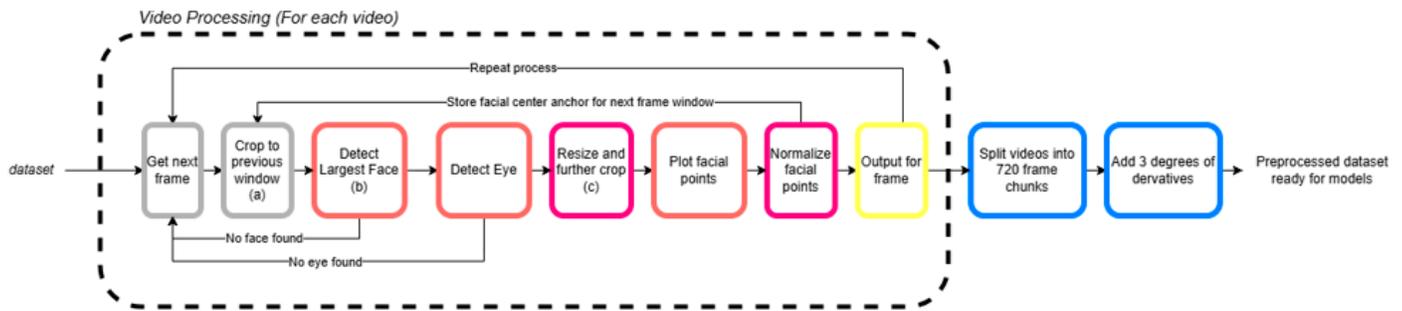

The DLib library used in the next step is sensitive to background noise; therefore, the face was further cropped to remove any background and resized to a uniform size (c). This cropped image was then passed into the DLib facial landmark predictor based on the pre-trained model *shape_predictor_68_face_landmark.dat* (Guo, 2019), a lightweight facial landmark model found in the popular DLib machine learning library (King, 2025). This predictor returned a series of sixty-eight Cartesian coordinate pairs marking the location of various facial features (Figure 2). This predictor identified 68 total points to extract from a face: seventeen from the chin, five left eyebrow, five right eyebrow, four nose bridge, five nose bottom, six left eye, six right eye, twelve outer lip, and eight inner lip. From this, the points were used to find the center of the face to aid in cropping the image for the next one. The face in the next frame would likely be in a similar position to the face in the current one. Finally, each frame's facial landmarks were appended to the dataset, recording the video identifier, the shape of the image, and the location of each facial feature. Once the video extraction steps finished, further data preprocessing steps were completed on the models. The final extraction steps result from adapting and refining the algorithm to best capture facial landmark data.

*Figure 2.*
*Facial Landmark Points, with Reference Facial Image and Without Reference Facial Image*

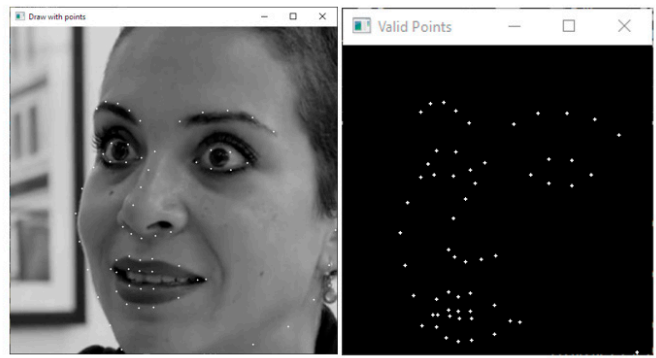

*Note.* By extrracting these facial landmark points, a model can train off these points directly and how they move, reducing the data required for training.

Initially, extracted pre-processed facial point coordinates fell into negative values, misidentifying faces, and jumping around to different faces in the scene. Negative values from the facial coordinates indicated there were facial landmarks outside the image. The predictor engine would then interpolate the locations of these landmarks. To handle these negative values correctly, each facial point was scaled to a uniform scale for each frame, removing the impact a negative value could have on a dataset. Next, the frontal face cascade classifier experienced issues with *hallucinations*, falsely identifying random objects as faces. This was solved using an eye detection cascade classifier to ensure that two eyes could be detected on the face that was identified by the earlier frontal face classifier. Lastly, multiple faces in one frame would sometimes confuse the facial detection causing a sudden shift in points. To solve this the researchers stored the previous face as a fixed point that could then be referred to. The face closest to that last fixed point would then be the one displayed in the output dataset.

*Data Preprocessing*

From the data extraction step, the points for each frame for each video were stored in a comma-separated (CSV) formatted file. Then, the data preprocessing component created a uniform dataset of videos with equal frame lengths. Before preprocessing, there were 362 authentic videos and 1861 artificially generated videos. These videos had frame lengths that ranged from less than 100 to over 1,500. The extracted dataset was 221,241 frames in length, with 362 videos for the real video dataset and 1,130,924 for the deepfake dataset.

After this, all video lengths were either padded or split to make a uniform distribution of 720 frames (30 seconds) among frame lengths throughout all videos in the dataset. Videos exceeding 720 frames were split into separate segments of 720 frames each. Then, videos above 600 frames were padded to 720 frames, while all videos less than 600 frames were dropped. Padding was applied by duplicating the last frame to simulate a static face for the remaining 720 frames. This ensured that each video was of the same length.

Lastly, the facial landmark columns were feature-engineered to obtain the first, second, and third differentials of the frame movement. The first differential captured the rate of change between consecutive frames, the second differential measured the acceleration of that change, and the third differential indicated how the acceleration evolved over time. This resulted in a dataset where each row corresponded to the 68 facial points of a specific frame in a video, along with a first, second, and third differential of each point's location.

The final dataset, standardized to 720 frames per video, consisted of 146,160 frames across 203 real videos and 723,600 frames across 1,003 deepfake videos.

## MODEL TESTING

The usefulness of this extracted and preprocessed dataset was tested in three different neural network architectures: an RNN, a CNN, and an ANN.

*RNN Model*

Developing this application involves creating a neural network capable of identifying patterns in facial landmark coordinates extracted from both deepfake and genuine videos. Recurrent neural networks (RNNs) are particularly suited to this task because they predict subsequent data points based on previously observed patterns in sequential data (Güera & Delp, 2018). By leveraging the trends uncovered in the facial landmarks, the RNN can effectively classify a given video's legitimacy.

The architecture of this model is mainly built to process the temporal patterns in the data to highlight the subtle inconsistencies in facial motion (see Figure 3). The data includes not only positional data but also utilizes the dataset's velocity, acceleration, and additional derivatives of point data to improve the model's ability to find deeper trends. Using TensorFlow and mathematical Python libraries such as Numpy and Pandas, all the frame data is homogenized into a single data frame with an additional column labeling the legitimacy of the frame. The rows are then resorted and regrouped by video index and fake label. The data is then split 70-15-15, where 70% of the data is used to train the model, 15% is used to validate, and the last 15% is used to test.

*Figure 3.*
*RNN Model Architecture*

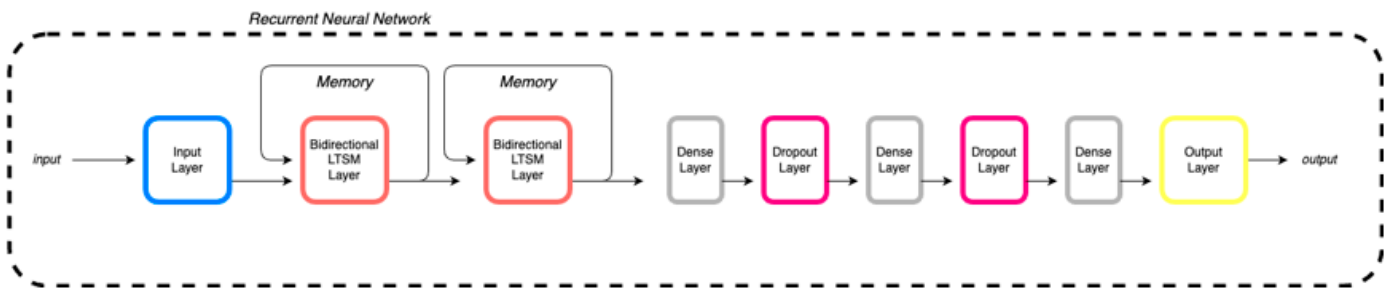

After the initial input layer, the middle layers of the RNN capture the temporal dependencies and patterns of the sequential data. Long short-term memory (LSTM) layers help retain long-term dependencies and prevent vanishing gradient problems. They work by deciding which information to discard from the previous cell of memory, figuring out which new information should be added to the memory, and which part of the memory should be output at the time step. The model also converges faster, meaning the weight adjustment reduces faster. After the RNN layers, the data flows through the dense layers. The dense layers extract higher-level features from the result of the RNN layers and make classification decisions. Dropout layers reduce overfitting by reducing the dependency of the model on any one neuron. The final dense layer is the probability score of the model, which exists between 0 and 1. The output layer then rounds the probability score to an integer and outputs 0 (real) or 1 (fake).

The model is compiled using the Adam optimizer. It is a fast and robust optimizer that supersedes *Momentum* and *RMSProp*. Adam updates weights based on the exponentially weighted average of past and squared gradients. The small learning rate of 0.0005 leads to careful and precise adjustments to the weights, which leverages the ability to process the complex dataset.

The quantitative results from the RNN model included a 96% accuracy, 94% precision, 100% recall, 97% F1-Score, and 1.0 ROC-AUC. Throughout training, it was observed that training and validation accuracy increased (see Figure 4). After epoch six, validation accuracy slightly leveled off, while training accuracy continued improving, which might indicate early signs of overfitting. In addition, model training and validation loss decreased as the model trained over each epoch (see Figure 5). Around epochs four to six, the validation loss stopped decreasing consistently and even slightly increased before going down again. This could indicate slight overfitting. By the final epochs, training loss was lower than validation loss, which was expected but should be monitored to avoid overfitting. Overall, the model was observed to be learning effectively.

*Figure 4.*
RNN Model Accuracy Over Training Epochs

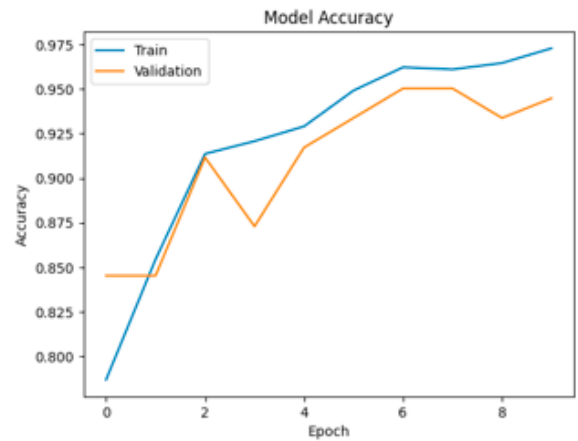

*Figure 5.*
RNN Model Loss Over Training Epochs

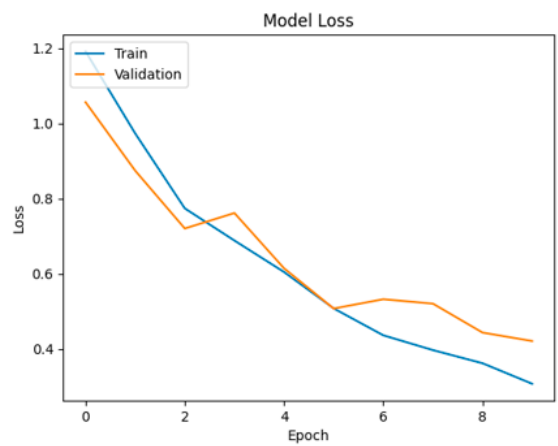

### CNN Model

Unlike the approaches by Güera and Delp (2018) and Ademiluyi (2023), the CNN model differs in using only facial feature extraction data. Most approaches follow a method similar to Güera and Delp (2018) and Ademiluyi (2023), utilizing CNNs to work directly on raw image data. To test the validity of using only the facial landmarks, the dataset is transformed into an image-like data structure to train on.

To do this, our dataset was split into one-second video splices. Then, for each facial point in the dataset, an image was created by "drawing a line" connecting the point coordinates as it progressed through that second of video. This then created a graph-like image that showed the movement of that single point. After this process was done to all 68 points, the images were combined into one multi-layered image of 68 layers. This image was then sent to the CNN model for training. Gaussian noise was also added to each input image to increase randomness and to prevent overfitting. Figure 6 shows this architecture and model.

*Figure 6.*
*Model of the CNN Architecture*

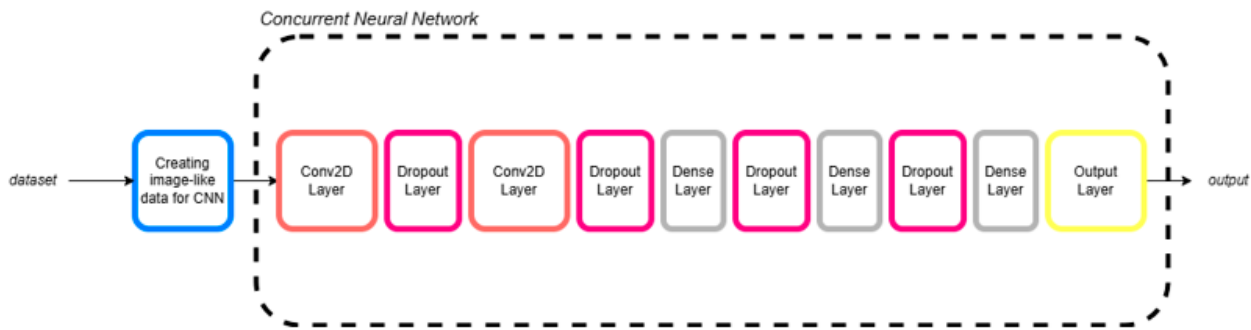

Next, the CNN model took in these images and used two 2D-convolutional layers, along with dropout layers, to focus on the patterns in the images. It then fed into a series of fully-connected layers for training. This model was trained via a cross-validation technique using three rounds with increasing epochs and batch sizes. This made the first round more impactful, while the latter rounds tended to be fine-tuned. The quantitative results from the CNN model included 77% accuracy, 78% precision, 72% recall, and a 77% F1-score. It was observed that the CNN model was highly sensitive to the number of epochs. With more epochs, the accuracy increased. However, due to the high computational load of the CNN model, a high number of epochs was prohibitive in computational power.

*ANN Model*

The ANN model served as a baseline for comparison with RNN and CNN models. The dataset was divided into one-second video chunks, where each sample contained 720 frames with the coordinate pairs of each facial point per frame. These frames were flattened into a single 97200-dimensional vector per sample before being processed by the ANN. The architecture consisted of five dense layers with dropout regularization, using ReLU activations to extract patterns. A final sigmoid activation layer predicted whether the input was real (1) or fake (0). The ANN was trained with binary cross-entropy loss and the Adam optimizer, providing an initial benchmark before testing more sequential architectures like RNNs and CNNs. The ANN model did much better than expected, with an accuracy of 93% after three rounds of cross-fold validation with incremental increases to the batch size. Furthermore, the quantitative findings included a 95% precision, 61% recall, 75% F1-score, and 0.80 ROC-AUC. The qualitative results showed that while accuracy and precision were comparable to the RNN, the ANN demonstrated a noticeably lower recall and F1-score (see Table 1). Additionally, it appeared to require more computational resources based on observed performance during testing.

## MODEL RESULTS

*Table 1.*
*Comparing Performance Results of the Models*

| Model | Metrics | | | |
|---|---|---|---|---|
| | Accuracy | Precision | Recall | F1 Score |
| RNN | 96% | 94% | 100% | 97% |
| ANN | 93% | 95% | 61% | 75% |
| CNN | 77% | 78% | 76% | 77% |

## CONCLUSION

Overall, the method of purely using facial landmarks for deepfake detection was effective, scoring high in both RNN and ANN. Our findings show that the feature extraction technique is best paired with the RNN, with a 96% accuracy and the lowest computational time among CNN and ANN. The feature extraction algorithm utilized 68 Cartesian coordinate pairs to extract facial features from both authentic and artificially generated videos. After applying data preprocessing techniques such as padding, resizing, and normalization, each model processed the 720-frame sequence and successfully performed binary classification, labeling each video as either real or deepfake. This extraction algorithm worked in various deep neural network architectures, promising a novel way of deepfake detection while suggesting high performance with decreased data load.

Future work could involve a detailed analysis of the models' computational times, parameter counts, and memory usage to enable a more comprehensive and rigorous comparison. This work could expand to include additional features besides facial characteristics in the future. For instance, using the same process on a full-body feature extraction may give similar results. Lastly, these models and feature extraction techniques could be integrated into a full-stack application, coupled with a convenient browser extension, to allow for real-time deepfake recognition in order to protect the users trusting this technology.

While our experiments on the Kaggle dataset demonstrated strong landmark-based detection performance, this curated data lacked many of the distortions found in real-world deepfakes, such as motion blur, varied lighting, and complex backgrounds typical of social media content. Future works could evaluate our RNN and ANN models on in-the-wild videos sourced from YouTube, TikTok, and publicly archived political ads. By introducing synthetic noise—such as Gaussian blur, resolution scaling, and temporal jitter—and testing across various frame rates and resolutions, the models could be adapted to handle more challenging, real-world conditions.

**ACKNOWLEDGMENT**

The researchers would like to thank Tejaswi Chintapalli for being our advisor for this project. She greatly encouraged us to keep going in the research process every time we encountered any challenges. Thank you!